\documentclass{article}
\usepackage{ijcai20}

\usepackage{times}

\usepackage{soul}
\usepackage[utf8]{inputenc}
\usepackage[small]{caption}
\usepackage{graphicx}
\usepackage{amsmath}
\usepackage{booktabs}
\usepackage{times}  
\usepackage{helvet} 
\usepackage{courier}  
\usepackage[hyphens]{url}  
\usepackage{anyfontsize}
\usepackage{amsmath}
\usepackage{amssymb}
\usepackage{graphicx}  
\usepackage{subcaption}
\usepackage{float}
\usepackage[hyperfootnotes=false,hidelinks=true]{hyperref}
\usepackage{relsize}
\usepackage{times}
\usepackage{helvet}
\usepackage{courier}
\usepackage{courier}
\usepackage{float}
\usepackage{graphicx}
\usepackage{subcaption}
\usepackage{amsfonts}
\usepackage{amsmath}
\usepackage{booktabs}
\usepackage[ruled,vlined]{algorithm2e}
\usepackage{algorithm, algorithmic}
\usepackage{color}
\usepackage{enumitem}
\usepackage{setspace}
\usepackage[bottom]{footmisc}
\usepackage{pifont}

\newcommand{\etc}{\emph{etc.}}
\newcommand{\UnnumberedFootnote}[1]{{\def\thefootnote{}\footnote{#1}
\addtocounter{footnote}{-1}}}

\title{Adversarial Graph Embeddings for Fair Influence Maximization over Social Networks}

\author{
Moein Khajehnejad$^{1}$$^{*}$\and
Ahmad Asgharian Rezaei$^{2}$$^{*}$\and
Mahmoudreza Babaei$^{3}$\and\\
Jessica Hoffmann$^{4}$\and
Mahdi Jalili$^{2}$\And
Adrian Weller$^{5,6}$\\\
\affiliations
$^{1}$Monash University\\
$^{2}$RMIT University \\
$^{3}$Max Planck Institute for Software Systems\\ 
$^{4}$The University of Texas at Austin\\
$^{5}$University of Cambridge \\ $^{6}$Alan Turing Institute \\
moein.khajehnejad@monash.edu, 
ahmad.asgharian.rezaei@rmit.edu.au,
babaei@mpi-sws.org, 
hoffmann@cs.utexas.edu,
mahdi.jalili@rmit.edu.au,
aw665@cam.ac.uk
}

\begin{document}

\maketitle
\UnnumberedFootnote{\scriptsize $^{*}$Authors contributed equally to this work.}

\begin{abstract}
Influence maximization is a widely studied topic in network science, where the aim is to reach the maximum possible number of nodes, while only targeting a small initial set of individuals. 
It has critical applications in many fields, including viral marketing, information propagation, news dissemination, and  vaccinations. 
However, the objective does not usually take into account whether the final set of influenced nodes is fair with respect to sensitive attributes, such as race or gender. Here we address \textit{fair influence maximization}, aiming to reach minorities more equitably. We introduce Adversarial Graph Embeddings: we co-train an auto-encoder for graph embedding and a discriminator to discern sensitive attributes. This leads to embeddings which are similarly distributed across sensitive attributes. We then find a good initial set by clustering the embeddings. We believe we are the first to use embeddings for the task of fair influence maximization. While there are typically trade-offs between fairness and influence maximization objectives, our experiments on synthetic and real-world datasets show that our approach dramatically reduces disparity while remaining competitive with state-of-the-art influence maximization methods.

\end{abstract}
\vspace{-4mm}

\section{Introduction}

A subfield of particular interest in network diffusion is influence maximization (IM), in which nodes and edges respectively represent individuals and connections. In its classical formulation, the goal is for the diffusion to reach the maximum number of nodes, while only disseminating the information to a few initial individuals, also called \textit{seeds}. 

Influence maximization is crucial in a variety of applications, including 
adopting new behavior in social networks \cite{richardson2002mining,kempe2003maximizing}, viral marketing \cite{babaei2013revenue}, propagation of information, disease spread \cite{banerjee2013diffusion,mirzasoleiman2012immunizing},
and social recommendations \cite{ye2012exploring}.  
However, \cite{kempe2003maximizing} showed that finding the best seeds is NP-hard. Numerous approximations and heuristics have been proposed. Several approaches identify the most influential nodes to maximize the total number of people influenced
\cite{li2018influence,kourtellis2013identifying}.
Using network structure, they consider central nodes (highest degree, closeness, betweenness, \etc) as the most influential nodes. Tackling the task from a different angle,
\cite{kempe2003maximizing} formulated the problem under the framework of discrete optimization. Their results significantly out-perform node selection heuristics, and showed that a provable approximate guarantee is obtainable. 
In this work, we 
adopt a fresh perspective: Adversarial Network Embeddings. We show our method is competitive with the previous influence maximization state-of-the-art, while  significantly improving \textit{fairness}.

Indeed,
our objective is \textit{fair influence maximization}, in which the resulting set of influenced nodes is diverse with respect to sensitive attributes, such as age, gender, race \etc. 
Adding fairness objectives to influence maximization yields an even harder task, which none of the previously mentioned techniques can tackle. However, these constraints are very relevant in today's society. For instance, consider the spread of a job opening, a loan advertisement, or even news. We want to make sure that belonging to a minority does not affect whether or not we would see this job opportunity or this loan offer. Moreover, as discussed in \cite{babaei2018purple}, receiving different types of news may be crucial to developing unbiased viewpoints. 
Making sure 
that everyone has a chance to see critical news, independently of the communities to which they belong, could be a stepping stone in the fight against fake news. \\

Fair influence maximization has been recently studied \cite{Ali2019OnTF,Tsang_dmt,fish_dmt}. 
However,
these methods suffer from a trade-off between fairness and the influence maximization objective. Using our Adversarial Network Embeddings, we achieve a highly diverse set of influenced nodes with respect to multiple attributes, while still achieving state-of-the-art influence maximization objectives on large synthetic and real-world networks.

\subsection{Our Contributions}
We highlight the following contributions:
\begin{enumerate}[leftmargin=*]
    \item We use an embedding approach to tackle 
    influence maximization.
    We believe this is the first time embeddings have been used for fair influence maximization.
    
    \item We introduce Adversarial Network Embeddings to address fair influence maximizaiton. Using an autoencoder coupled with a discriminator in an adversarial setting, we obtain embeddings which are similarly distributed across sensitive attributes.
    \item Our approach achieves state-of-the-art results for influence maximization (comparing to the greedy approach, previous state-of-the-art) in experiments on synthetic and real-world datasets.
    \item Our method also achieves high diversity in the context of fair influence maximization.
\end{enumerate}
Our work provides a fresh tool that we hope could be helpful to other social network settings where fairness is relevant, such as fair clustering or fair node-level classification.

\section{Related Work}
Adversarial approaches to fairness were introduced recently \cite{madras,adel}. To our knowledge, only a few studies have considered fairness in the context of network diffusion. In this section, we review 
work on influence maximization, node embedding approaches, and recent 
investigations of fair influence maximization. 
\subsection{Influence Maximization}
Influence maximization was first introduced as an algorithmic problem by \cite{richardson2002mining} by proposing a heuristic approach to find an initial set of nodes to maximum the number of further adopters. Over the years, extensive research 
has focused on cascading behavior, diffusion and spreading of ideas, information or diseases, by identifying a set of initial nodes which maximizes the influence through a network \cite{leskovec2007cost,kempe2003maximizing,richardson2002mining,leskovec2010inferring}.  
  
Identifying individuals who are good initial seeds for spreading of information is studied in two ways: (i) find the set of most central nodes based on network structural properties \cite{kourtellis2013identifying,kempe2003maximizing}; or (ii) tackling the problem as discrete optimization \cite{goyal2013minimizing,kempe2003maximizing,babaei2013revenue}. 

\cite{kempe2003maximizing} studied influence maximization under different social contagion models such as Linear Threshold (LT) and Independent Cascade (IC) models. They showed that although finding the optimal solution is NP-hard, submodularity of the influence function can be used to obtain provable approximation guarantees. Since then, there has been a large body of work studying various extensions \cite{goyal2013minimizing,carnes2007maximizing} among which \cite{keikha2020influence} takes advantage of a network embedding approach. 

\subsection{Network Embedding}
Learning a low-dimensional embedding of the nodes in the network is at the core of our proposed approach. 
Generally, the network embedding problem proposes to map nodes to a low dimensional space such that the network structure can be reconstructed. 
Network embeddings have proven their efficiency in classification and clustering problems \cite{hamilton2017representation,cao2015grarep}, and have attracted much attention from the machine learning and data mining communities \cite{cai2018comprehensive,hamilton2017representation,tang2015line}. 
Consequently, several methods have been proposed based on random-walk based models 
\cite{khajehnejad2019simnet,grover2016node2vec}, deep learning architecture \cite{wang2016structural}, and graph neural networks \cite{hamilton2017inductive}.
However, to our knowledge no one has considered network embedding to address the problem of fair influence maximization.

\subsection{Contemporary Works}
Only a few recent works consider fairness in influence maximization.
Recently, several works promote diversity in choosing the seeds 
\cite{benabbou2018diversity,aghaei2019learning}. \cite{babaei2016efficiency} also show that users are sub-optimal in selecting their sources in social media in the sense of receiving 
diverse information. 
But these works do not consider a fairness criterion. The notion of individual fairness (similar individuals should be treated similarly) and group fairness (on average, members of different groups are treated similarly) for 
influence maximization were proposed by \cite{fish_dmt} and \cite{Tsang_dmt}, respectively.  

These two studies are the most related to our work. However,
they suffer from a trade-off between the objectives of influence maximization and fairness. This trade-off becomes worse as the number of sensitive attributes increases.
Our method achieves state-of-the-art influence maximization results, even with multiple sensitive attributes, while also increasing fairness. We next present how our Adversarial Network Embeddings achieve these desired properties.

\section{Methodology}
\begin{figure*}[h!]
    \centering 
    \includegraphics[width=0.8\textwidth,height=7cm]{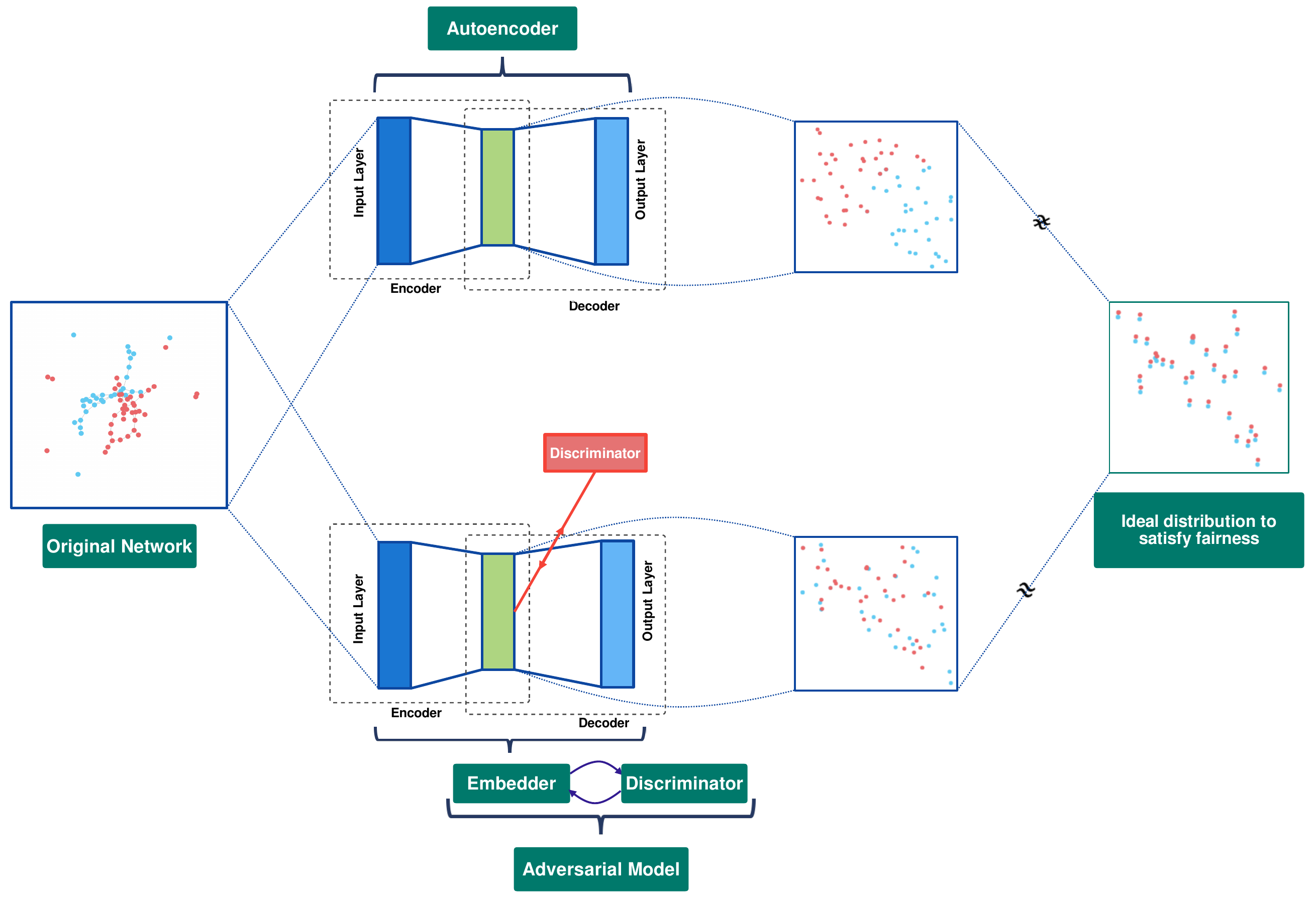}
    \caption{\small Illustrating the comparison between an auto-encoder and an adversarial system consisting of an auto-encoder interacting with a discriminator. The method is applied on a synthetic graph and the node representations in the new space are compared with an ideal case of equal distribution for both communities. The adversarial setting (bottom pathway) leads to a much more similar distribution to the ideal case.}\label{Fig1}
     \vspace{-4.5mm}
\end{figure*}
In the \textit{fair influence maximization} problem, we aim to influence the maximum number of nodes, while ensuring that the fraction of influenced nodes is (approximately) equal across predefined groups (e.g. minorities). For ease of exposition, we assume in this section that there are only two groups $A$ and $B$, but the results extend naturally to more groups. Let $I_A$ and $I_B$ be the expected total number of influenced nodes belonging to groups $A$ and $B$, respectively. We say a spreading process is \emph{fair} if, by the end of it, we have:
$$ \frac{I_A}{|A|} \approx \frac{I_B}{|B|}$$ 

We now provide the intuition and motivation behind our embedding approach. If we pick $S$ as our set of initial seeds, let $q_S(U)$ be the probability that the set of nodes $U$ was influenced by the end of the spreading process. We also write $U_A$ (resp. $U_B$) for the set of nodes of $U$ which belong to $A$ (resp. to $B$), and $\mathcal{P}(S)$ the set of all subsets of a set $S$. To compute the expected fraction of influenced nodes in $A$, we sum over all possible disjoint infection patterns:
\begin{align*}
I_A &= \sum_{u_a \in \mathcal{P}(A)} \mathbb{P}(u_a) \cdot |u_a|
\end{align*}
The probability that all nodes of $u_a$ are infected is given by marginalizing over all subsets of $B$: $\mathbb{P}(u_a) = \sum_{u_b \in \mathcal{P}(B)} q (u_a + u_b)$. Therefore:
\begin{align*}
I_A &= \sum_{u_a \in \mathcal{P}(A)} \sum_{u_b \in \mathcal{P}(B)} q (u_a + u_b) \cdot |u_a| 
\end{align*}
Reindexing, and dividing by $|A|$:
\begin{align*}
\frac{I_A}{|A|} &= \frac{1}{|A|} \cdot\sum_{U \in \mathcal{P}(\{1, \dots, N\})} q_S(U) \cdot | U_A | 
\end{align*}

This is where our Adversarial Network Embedding becomes useful. If we knew that for the sets $U$ of high probability mass ($i.e.$ the most likely sets of influenced nodes, with $q_S(U)$ of highest values), we had $\frac{| U_A |}{| A|} \approx \frac{| U_B |}{|B|}$, then we could claim that $ \frac{I_A}{|A|} \approx  \frac{I_B}{|B|}$. Intuitively, by matching the distributions of nodes from $A$ and $B$ in the embedding space (to have $\frac{| U_A |}{| A|} \approx \frac{| U_A |}{|B|}$ for as many $U$ sets as possible), and picking seeds in densely populated areas (so as to put more probability mass where the above equality is verified), we design our 
framework, which we describe next. 

To achieve the desired adversarial network embedding for an input network with two communities, inspired by 
\emph{Generative Adversarial Networks} \cite{goodfellow2014generative}, we design an adversarial setting in which the embedder plays against a discriminator. In our setting, the discriminator distinguishes between the embeddings of the two communities. Concurrently the embedder tries to generate embeddings that are indistinguishable by the discriminator. In other words, the discriminator forces the embedder to generate embeddings for the two communities that are coming from distributions which are as similar as possible. We train the embedder and the discriminator with the following coupled loss functions:
\begin{align*}
\begin{cases}
   \mathcal{L}_{E} =  \mathcal{L}_{recon}   - \beta\mathcal{L}_{\mathcal{D}} & \\
   \mathcal{L}_{D}= \mathcal{D}(Z_A)-\mathcal{D}(Z_B) & Z_A= \textit{E}(X_A), Z_B= \textit{E}(X_B)
  \end{cases}
\end{align*}
where $X_A$ , $X_B$ refer to the vector representation of nodes in communities $A, B$, and $Z_A, Z_B$ are their corresponding vector representation in the embedding space. $E$ and $L_E$ represent the embedder function and its relative loss function. $L_{recon}$ denotes the reconstruction loss of a standard auto-encoder. Similarly, $D$ and $L_D$ refer to the discriminator function and its loss. The discriminator function computes the distance between the distribution of the given embeddings with the initial distribution of the embeddings of nodes of community A.
Figure \ref{Fig1} depicts the whole process of our method in a graphical way. Details are shown in Algorithm \ref{alg:advesarial-alg}. We can further extend the above setting in order to address fairness in influence maximization with respect to multiple sensitive attributes. In this case, we want our embeddings to have similar
\begin{algorithm}[h]
\caption{\textsc{Adversarial Graph Embedding}: Adversarially trains the \textit{Embedder} function $E$ with a \textit{Discriminator} function $\mathcal{D}$.}
  \begin{algorithmic}[1]
    \REQUIRE Input network ($X$) with two communities ($X_A, X_B$), \textit{Embedder}~($E$) and \textit{Discriminator}~($\mathcal{D}$) functions, with parameters ($\theta_E, \theta_{\mathcal{D}}$).
      \STATE Pre-train $E$ on $X$ with $\mathcal{L}_{recon}(X) = \vert\vert X - \hat{X} \vert\vert^2_2$.
      \STATE $Z_A \gets E(X_A), Z_B \gets E(X_B)$
      \STATE Pre-train $\mathcal{D}$ on $Z_A, Z_B$ with $\mathcal{L}_\mathcal{D}(Z) = \mathcal{D}(Z_A) - \mathcal{D}(Z_B)$
      
      \FOR{$epoch \in \{1, \dots, n\}$} 
           \FOR{$X^{r} \in X$}
                \STATE $\theta_E = \theta_E - \nabla_{E} [\mathcal{L}_{recon}(X^r) - \beta \mathcal{D}(E(X^r_A)) + \beta \mathcal{D}(E(X^r_B))]$
                \STATE $Z^r_A \gets E(X^r_A), Z^r_B \gets E(X^r_B)$
                \STATE $\theta_{\mathcal{D}} = \theta_{\mathcal{D}} - \nabla_{\mathcal{D}} [\mathcal{D}(Z^r_A) - \mathcal{D}(Z^r_B)]$
            \ENDFOR
      \ENDFOR
      
      \STATE \mbox{\bf Return} $\theta_E, \theta_{\mathcal{D}}$
  \end{algorithmic}
  \label{alg:advesarial-alg}
\end{algorithm}
distributions for each value of our different sensitive attributes. To do so, we add one discriminator and one extra term to the embedder loss per sensitive attribute. Then during the adversarial training, the embedder plays against a set of discriminators (one discriminator per attribute). For an instance of two sensitive attributes, the embedder and the discriminator's loss functions would be as follows:
\begin{align*}
\begin{cases}
   \mathcal{L}_{E} =  \mathcal{L}_{recon}   - \beta_1\mathcal{L}_{\mathcal{D}_{1}} - \beta_2\mathcal{L}_{\mathcal{D}_{2}}& \\
   \mathcal{L}_{\mathcal{D}_{1}}= \mathcal{D}_1(Z_{A_{1}})-\mathcal{D}_1(Z_{B_{1}})& \resizebox{0.37\hsize}{!}{$Z_{A_{1}}= \textit{E}(X_{A_{1}}), Z_{B_{1}}= \textit{E}(X_{B_{1}})$} \\
    \mathcal{L}_{\mathcal{D}_{2}}= \mathcal{D}_2(Z_{A_{2}})-\mathcal{D}_2(Z_{B_{2}})& \resizebox{0.37\hsize}{!}{$Z_{A_{2}}= \textit{E}(X_{A_{2}}), Z_{B_{2}}= \textit{E}(X_{B_{2}})$} \\
\end{cases}
\end{align*}

After reaching a favorable low-dimensional representation of the network where nodes of different communities are distributed similarly in the embedding space, the final step is to choose the proper seeds. We can safely perform the selection in the embedding space since our embedding method is invertible (the original graph can be reconstructed from the embeddings).
For the case of one sensitive attribute, our goal is to choose an initial set of influential seeds $S$. We examine the two following approaches:

\begin{itemize}
\item (i) Normal Selection: This applies a $k$-means method on the $Z$ space with $k = |S|$ to select the resulting $k$ cluster centroids as initial seeds. 

\item (ii) Fair Selection: In normal selection (above), depending on the network structure, seeds might come from just one community, leading to disparity. To tackle this concern, we propose an alternative method introduced in Algorithm \ref{alg:FS}. Assume we want to select $|S| = k \times s$ nodes from $Z$ as the initially influenced nodes. We start by performing a $k$-means algorithm to group all nodes in $Z$ into $k$ clusters. Then, in each cluster, we select the $s$ nearest neighbours to the centroid and determine whether they are members of community $A$ ($N_A$) or $B$ ($N_B$). 
We also divide all the nodes of each cluster into two sub-clusters with nodes belonging to $A$ or $B$. Finally, we exploit the $k$-means algorithm on each of these sub-clusters using $k = |N_A|$ and $k = |N_B|$ respectively and obtain the resulting centroids to have selected $s = |N_A|+ |N_B|$ seeds from each of the initial $k$ clusters giving us $|S| = k \times s$ seeds from the whole network.
\end{itemize}

\begin{figure}[b]
\centering
\hspace{1cm}
\begin{subfigure}[h!]{0.2\linewidth}
\centering
    \vfill
    \makebox[0.5\linewidth][c]{\includegraphics[scale=0.16]{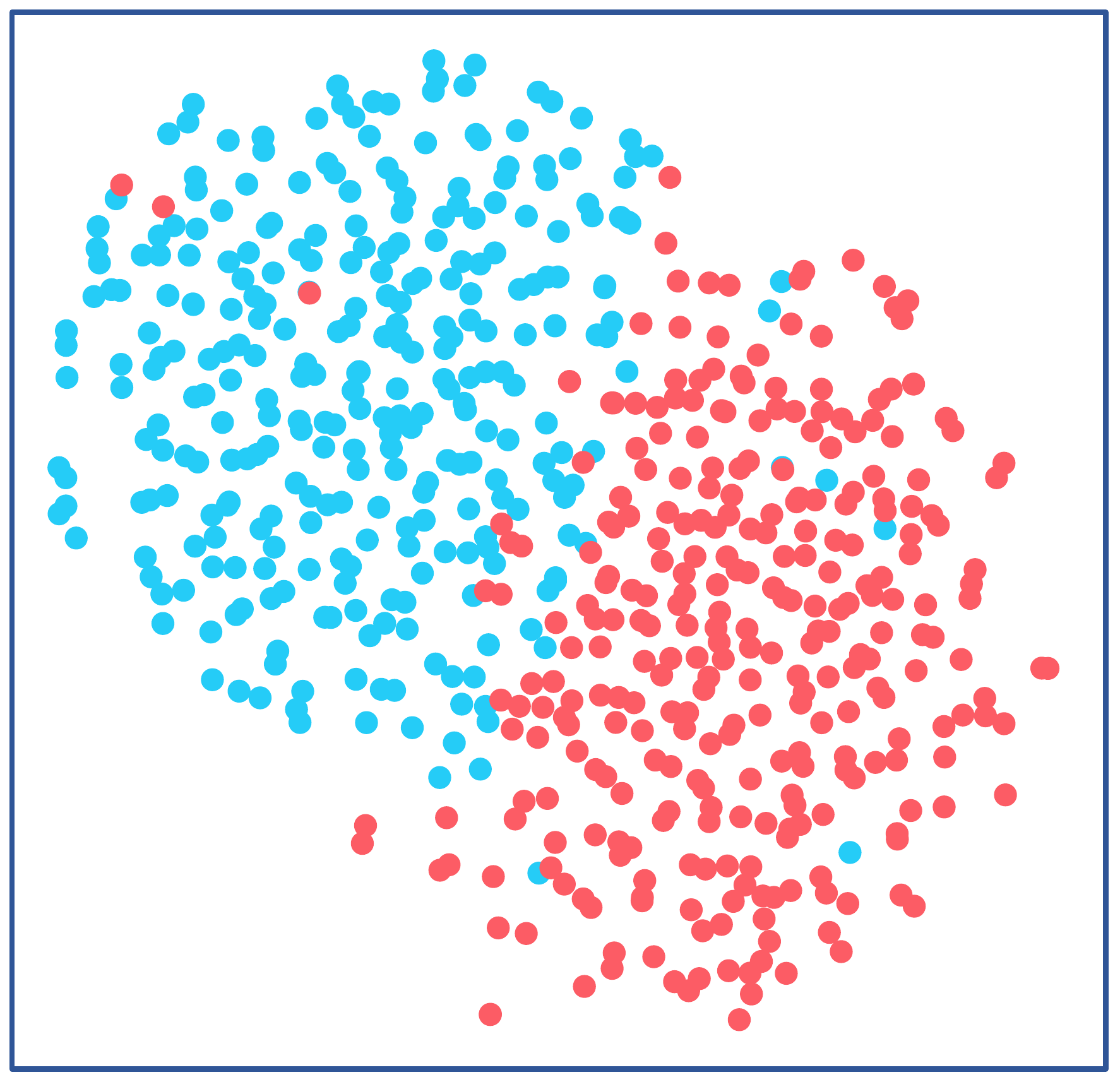}}
  \captionsetup{justification=centering}
   \caption{\small Standard auto-encoder} 
    \label{fig2:SynLargePre}
\end{subfigure}
\hspace{2cm}
\begin{subfigure}[h!]{0.4\linewidth}
\centering
    \vfill
    \makebox[0.5\linewidth][c]{\includegraphics[scale=0.16]{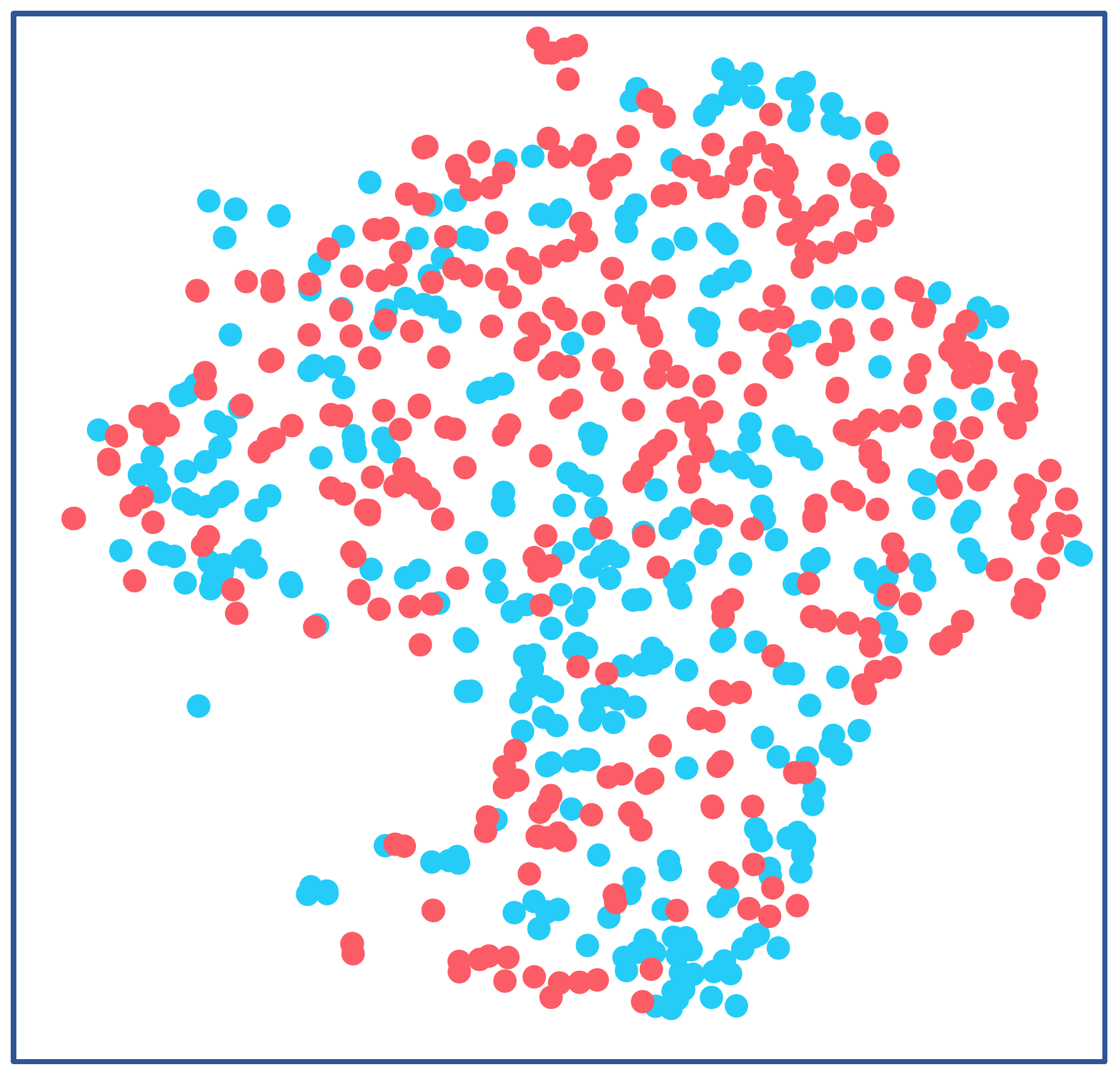}}
   \captionsetup{justification=centering}
    \caption{\small Adversarially \\ trained auto-encoder} 
    \label{fig2:SynlargeFair}
\end{subfigure}
\vspace{-0.25cm}
\caption{\small Embedding space for the standard and adversarially trained auto-encoders. Nodes of different communities are mapped to two segregated sub-spaces in (a), while being similarly distributed in the embedding space of (b).}
\label{Fig2}
\vspace{-1.5mm}
\end{figure}

We now show the proposed method's effectiveness using a small synthetic data set. Figure \ref{Fig1} illustrates the high performance and efficiency of the system in re-creating the network using our model. A standard auto-encoder generates a network where the two communities are completely distinguishable which is representative of completely different distributions. However, with our Adversarial Graph Embeddings, the two distributions mostly overlap - it is hardly possible to separate the two communities. The results are even more striking in Figure \ref{Fig2} which shows the embedding space of our proposed method in comparison with the embedding space of a standard auto-encoder for a large synthetic data set. It can be seen that using our algorithm, nodes from different communities show similar distributions in the embedding space.


\begin{algorithm}[t]
\caption{\textsc{Fair Selection}: Selects $|S|$ initial seeds attempting to choose the most influential ones.}
  \begin{algorithmic}[1]
    \REQUIRE Number of clusters $k = |S|$,  Embedding space $Z$
      \STATE $\{ g^i\}_{i=1}^{k} \gets \textrm{\textsc{ClustersCenters}}(k,Z)$
      \STATE $\{ G^i\}_{i=1}^{k} \gets \textrm{\textsc{ClustersPoints}}(k,Z)$
      \STATE $s \gets \frac{|S|}{k}$
      \FOR{$i \in \{1, \dots, k\}$}
           \STATE $N^i \gets \textrm{KNN}(s,g^i,G^i)$
           \STATE $N^i_A \cup N^i_B \gets N^i$ 
           \STATE $G^i_A \cup G^i_B \gets G^i$
           \STATE $S_0 \gets \textrm{\textsc{ClustersCenters}}(|N^{i}_{A}|,G^i_A) \cup \textrm{\textsc{ClustersCenters}}(|N^{i}_{B}|,G^i_B)$
      \ENDFOR
      \STATE \mbox{\bf Return} $S_0$
  \end{algorithmic}
  \label{alg:FS}
\end{algorithm}
In Algorithm \ref{alg:FS}, the function \textsc{ClustersCenters}($k,Z$) gives the set of $k$ centroids when performing $k$-means on the $Z$ space and \textsc{ClustersPoints}($k,Z$) returns the corresponding $k$ cluster of points. Finally, KNN($s,g^i,G^i$) outputs $s$ nearest neighbours of $g^i$ in the space of $G^i$. 
\section{Dataset}
\begin{figure*}[h!]
\centering
\vfill
    \hspace{-2mm}
    \begin{subfigure}[h!]{0.24\textwidth}
    \centering
   \vfill
        \makebox[0.45\linewidth][c]{\includegraphics[width=\textwidth]{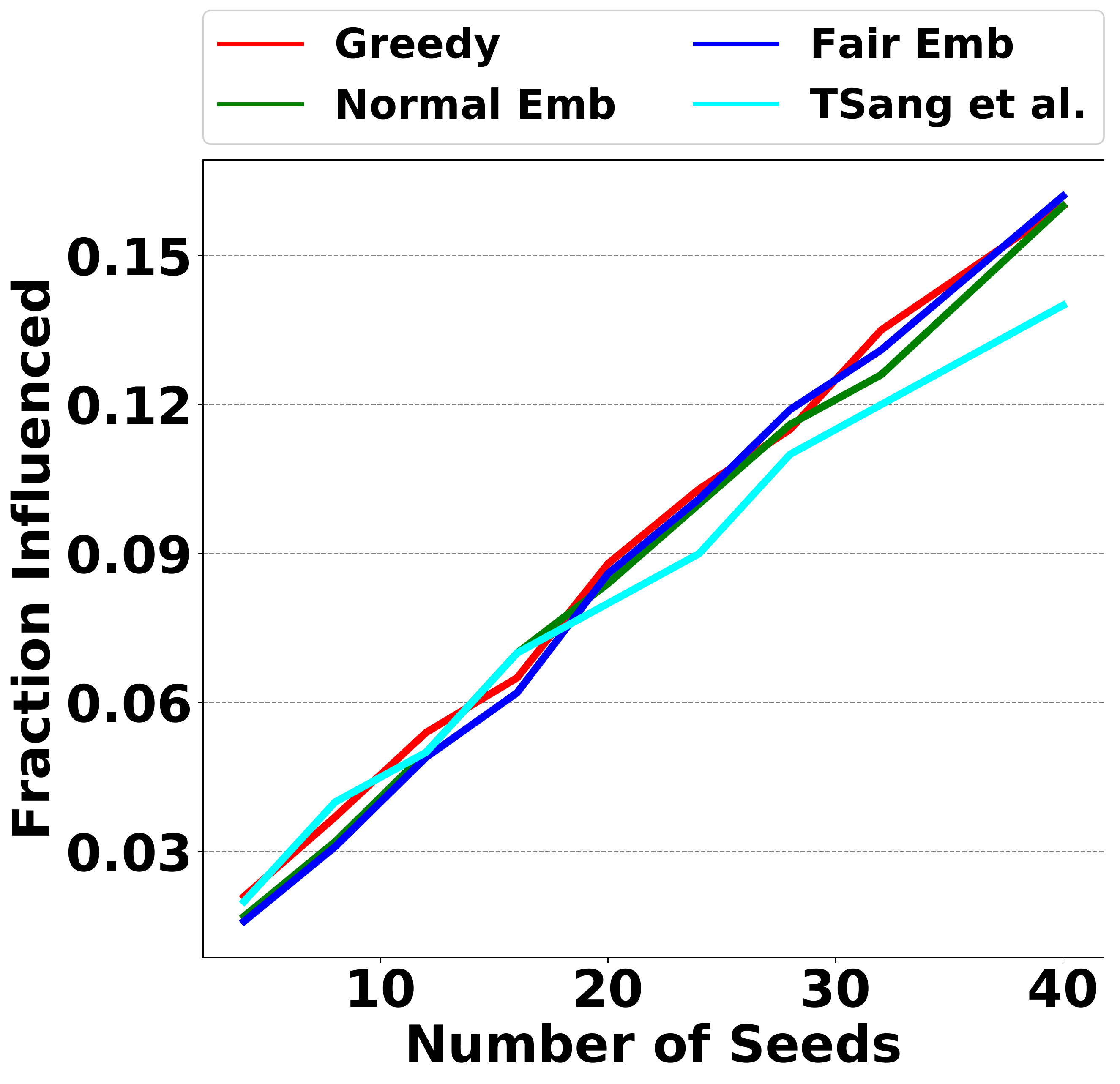}}
        \captionsetup{justification=centering}
        \caption{\small The total fraction of the influenced nodes}
    \end{subfigure}
    \hspace{-0.2mm}
    \begin{subfigure}[h!]{0.24\textwidth}
    \centering
        \makebox[0.45\linewidth][c]{\includegraphics[width=\textwidth]{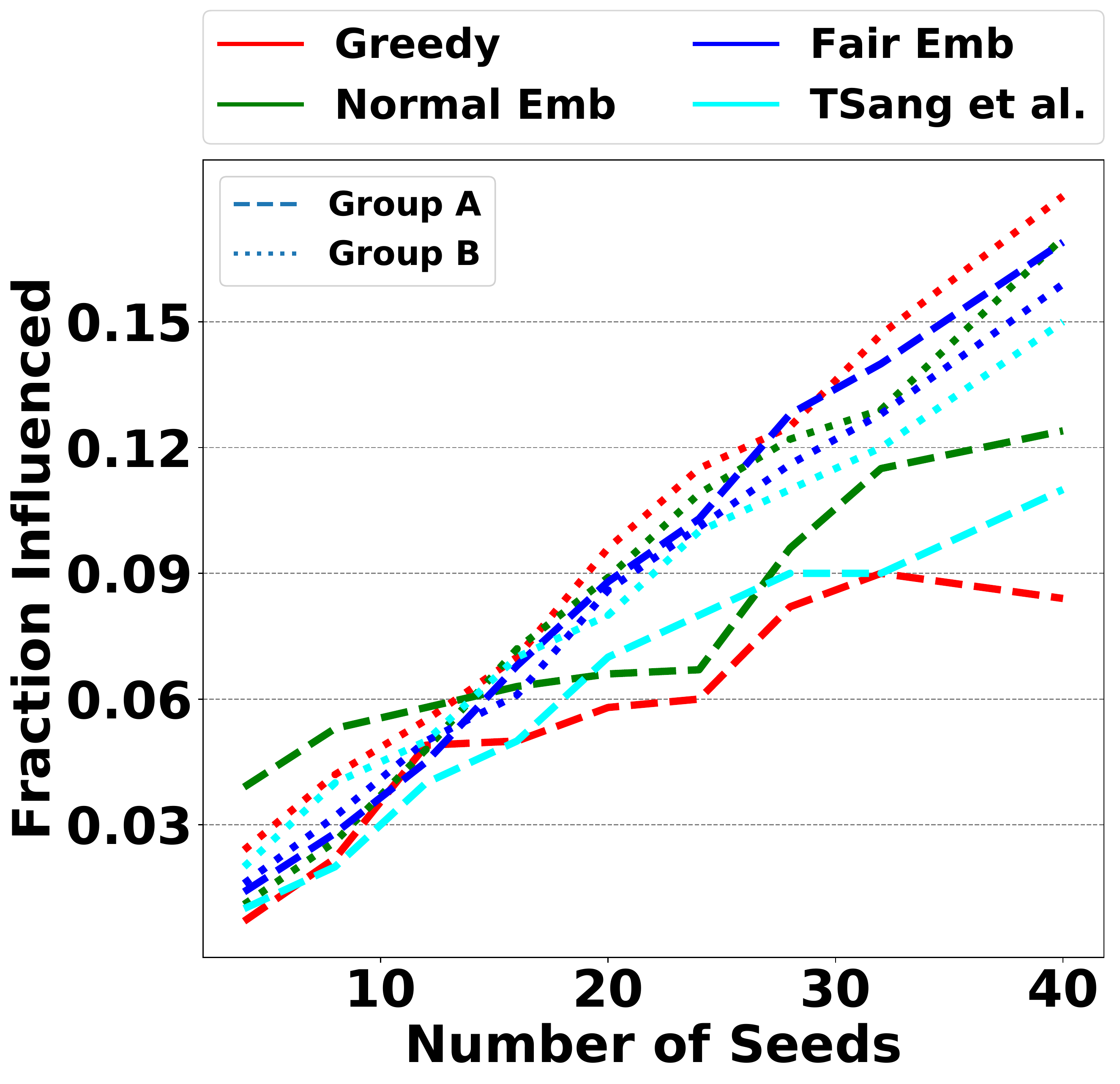}}
        \captionsetup{justification=centering}
        \caption{\small The fractions of influenced nodes in the two groups}
    \end{subfigure}
    \hspace{0.5mm}
    \begin{subfigure}[h!]{0.24\textwidth}
    \centering
    \vfill
        \makebox[0.45\linewidth][c]{\includegraphics[width=\textwidth]{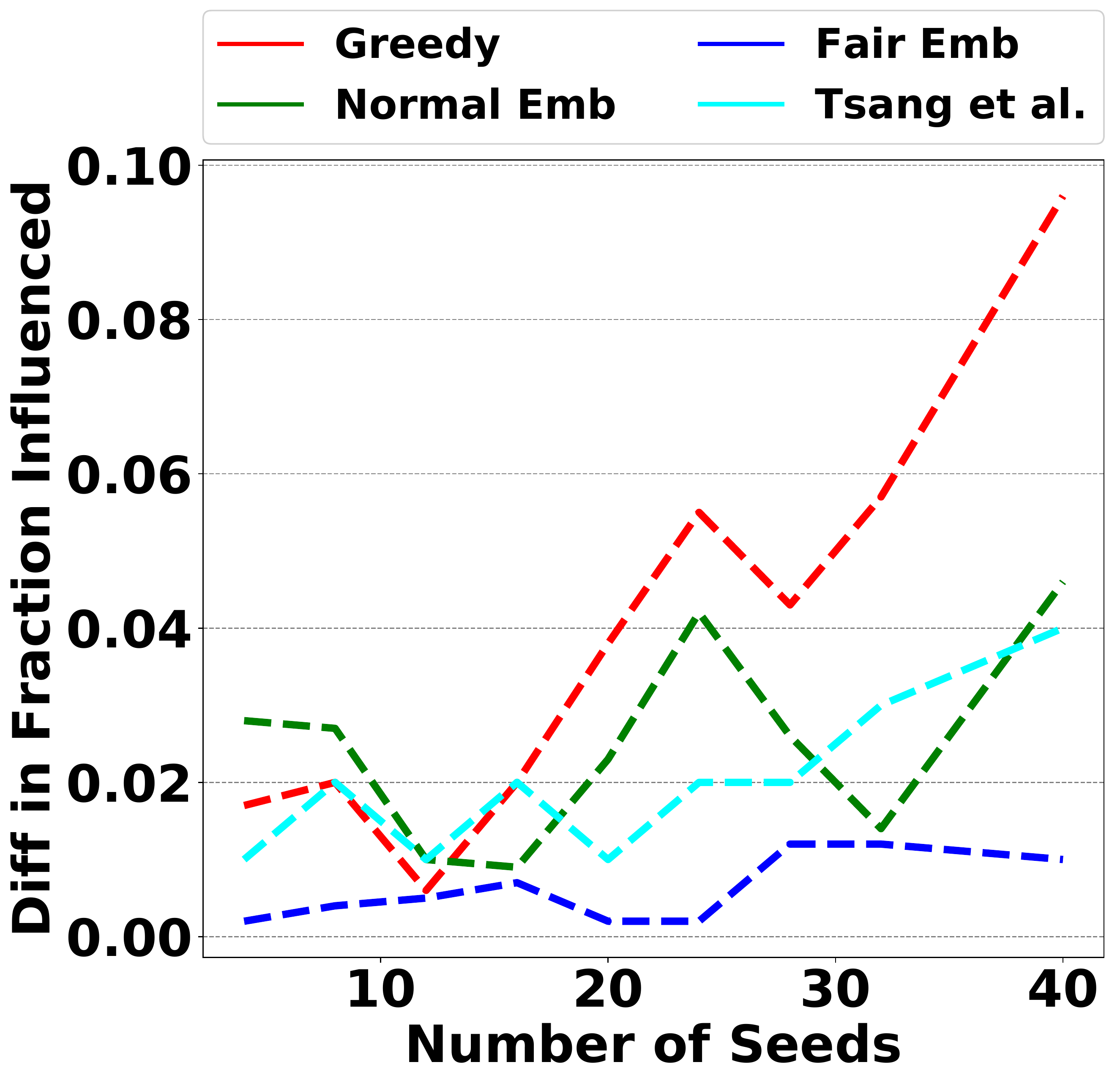}}
        \captionsetup{justification=centering}
        \caption{\small The difference in influenced nodes fractions in the two groups}
    \end{subfigure}
    \hspace{0.5mm}
    \begin{subfigure}[h!]{0.24\textwidth}
    \vspace{-0.3mm}
    \centering
    \vfill
        \makebox[0.45\linewidth][c]{\includegraphics[width=1.32\textwidth,height=4.35cm]{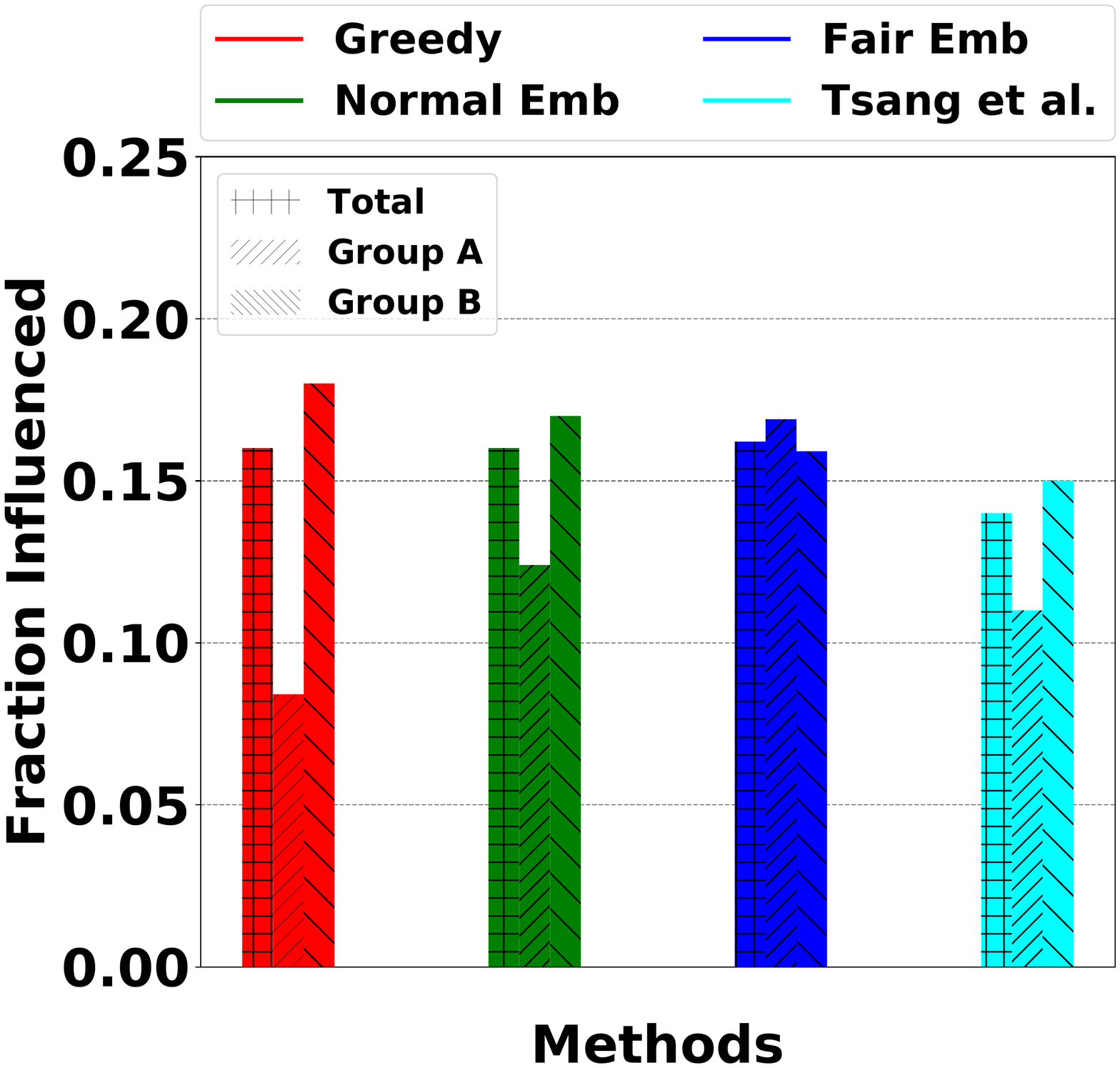}}
        \captionsetup{justification=centering}
        \caption{\small Total and group influence for 40 initial seeds}
    \end{subfigure}
\vspace{-2.5mm}
\caption{\small  Comparison of the results of the \emph{Normal Embedding} (green) and \emph{Fair Embedding} method (blue), as well as the \emph{Greedy} method (red), and \emph{Tsang et al.} (cyan) as baselines over the Rice-Facebook dataset. It can be seen that the \emph{Fair Embedding} approach influences a fair fraction of both groups while it does not affect the total influence. 
}
\label{Fig3:Result_real}
\vspace{-2.5mm}

\end{figure*}

\begin{figure*}[h!]
\centering
\hspace{-3mm}
    \begin{subfigure}[h!]{0.24\textwidth}
       \centering
        \vfill
        \makebox[0.5\linewidth][c]{\includegraphics[width=\textwidth]{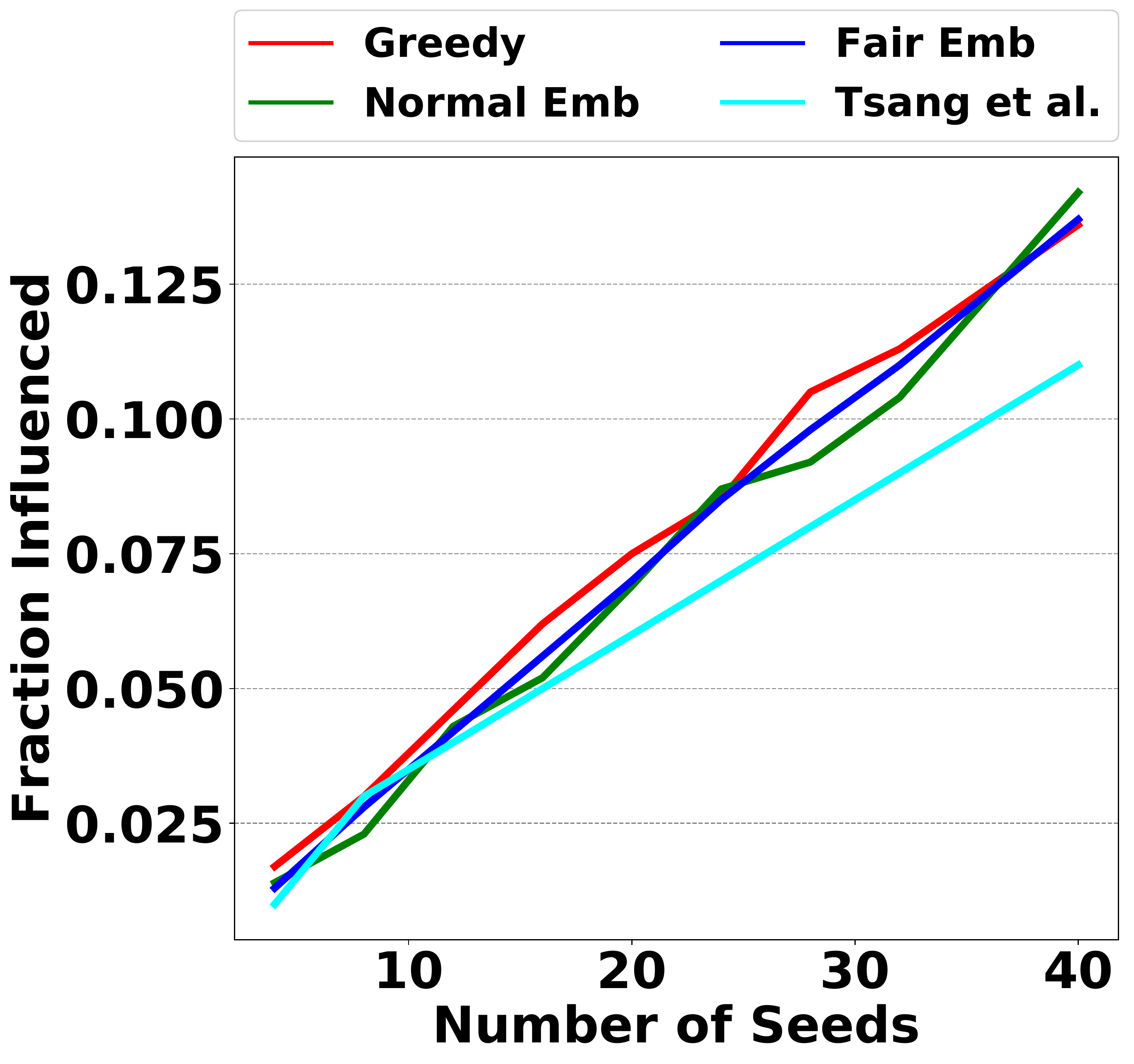}}
        \captionsetup{justification=centering}
        \caption{ \small The total fraction of the influenced nodes}
    \end{subfigure}
    \hspace{0.5mm}
    \begin{subfigure}[h!]{0.24\textwidth}
    \centering
         \vfill
       \makebox[0.5\linewidth][c]{\includegraphics[width=\textwidth]{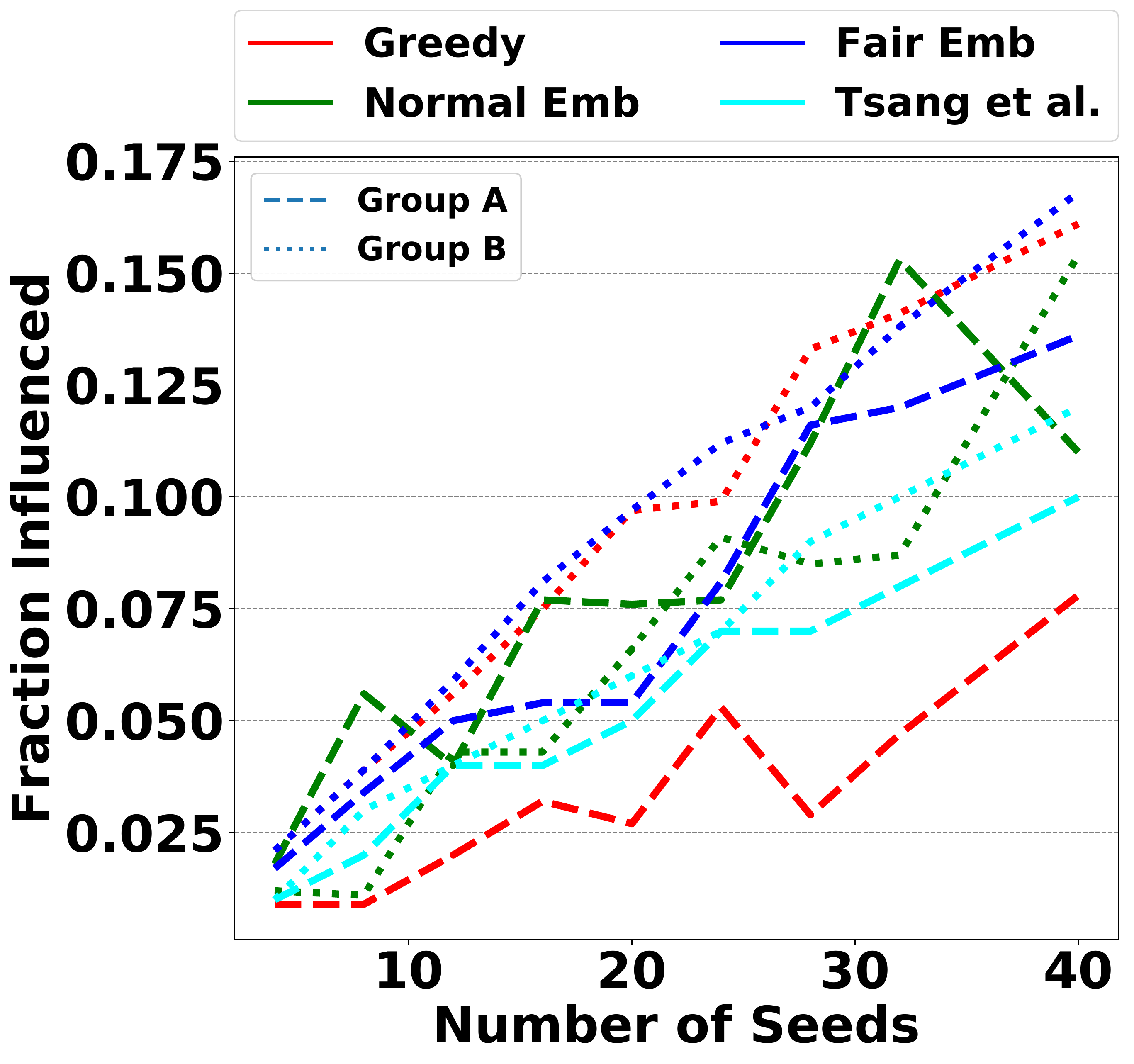}}
        \captionsetup{justification=centering}
        \caption{\small The fractions of influenced nodes in the two groups}
    \end{subfigure}
    \hspace{0.5mm}
    \begin{subfigure}[h!]{0.24\textwidth}
    \centering
        \vfill
        \makebox[0.5\linewidth][c]{\includegraphics[width=\textwidth]{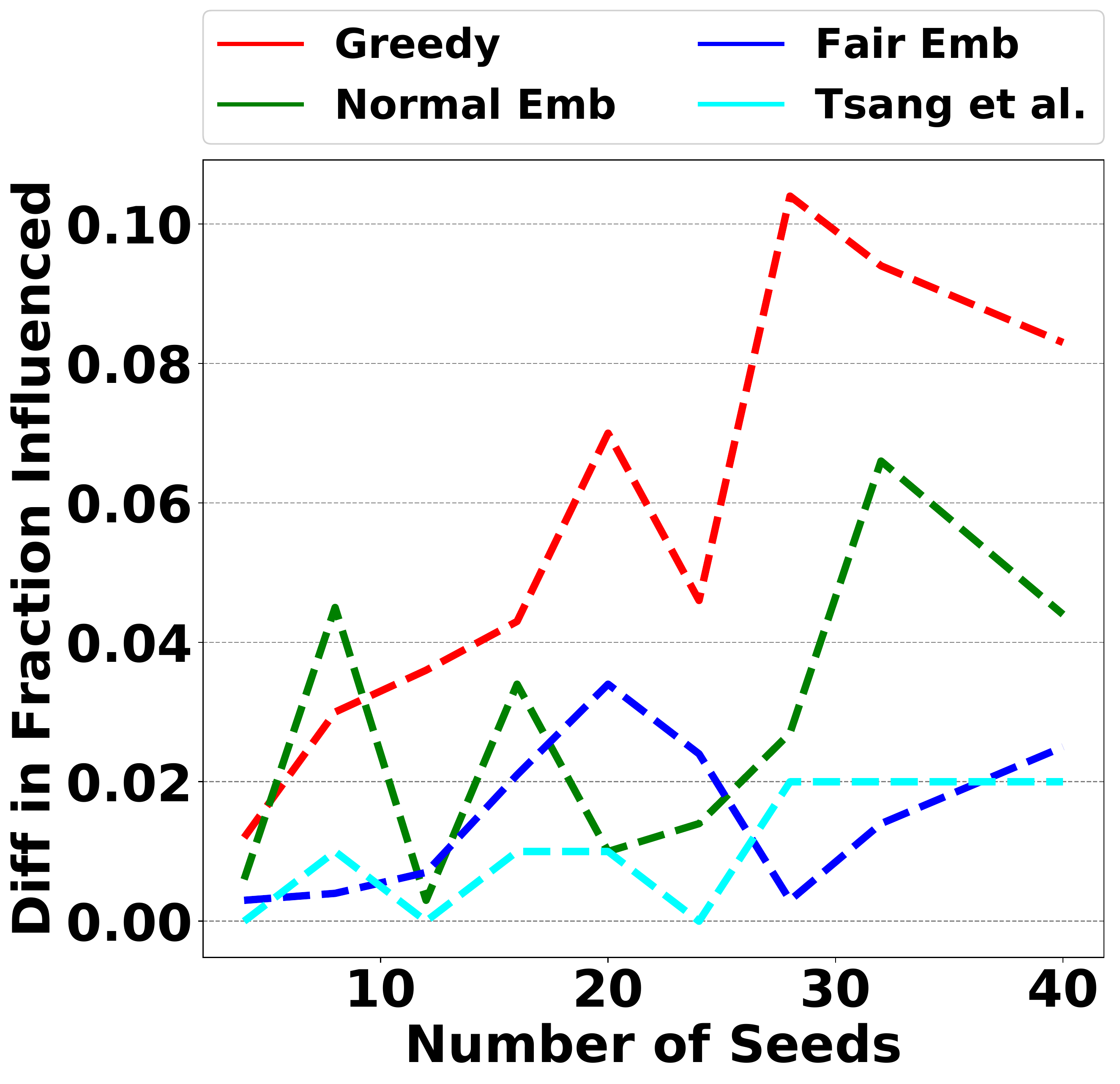}}
        \captionsetup{justification=centering}
        \caption{\small The difference in influenced nodes fractions in the two groups}
    \end{subfigure}
    \hspace{0.5mm}
        \begin{subfigure}[h!]{0.24\textwidth}
        \vspace{-0.3mm}
        \centering
        \vfill
        \makebox[0.5\linewidth][c]{\includegraphics[width=1.45\textwidth,height=4.35cm]{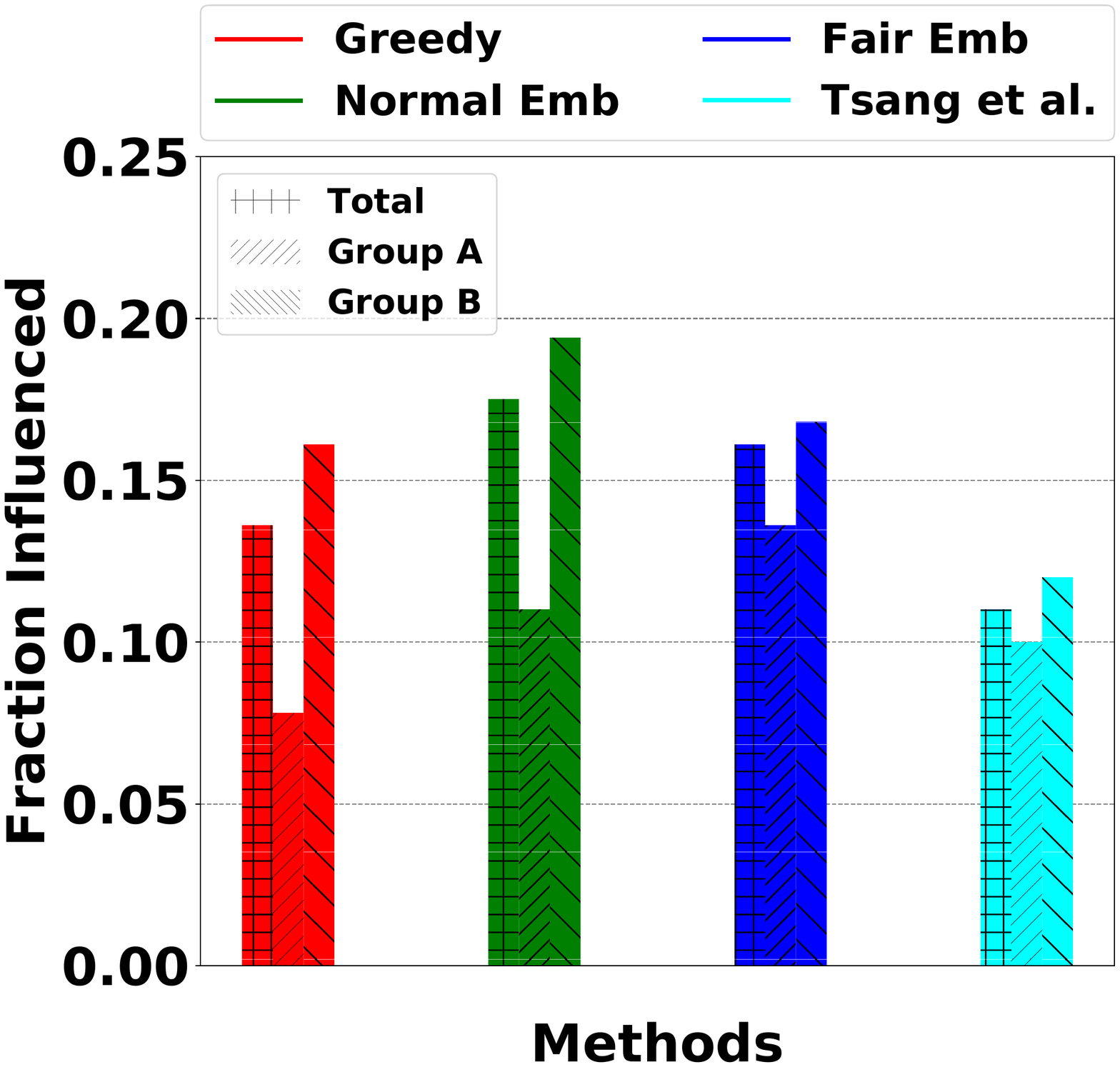}}
        \captionsetup{justification=centering}
        \caption{\small Total and group influence for 40 initial seeds}
    \end{subfigure}
\vspace{-2.5mm}
\caption{\small Comparison of the results on the synthetic dataset. Similarly, the \emph{Fair Embedding} (blue) method outperforms the \emph{Normal Embedding} (green), \emph{Tsang et al.} (cyan), and \emph{Greedy} (red) methods in enhancing the fairness while being very close to them in terms of total influence. 
}
\label{Fig4:Result_synth}
\vspace{-3.5mm}
\end{figure*}

We evaluate the performance of 
our approach on real and synthetic data.\footnote{\scriptsize {For code and details of parameter tunings please refer to: \href{https://github.com/Ahmadreza401/fair_influen_max}{https://github.com/Ahmadreza401/fair\_influen\_max}}}

\subsection{Real Dataset}
\textbf{Dataset Description}: The real dataset is the \emph{Rice University Facebook} dataset, collected by \cite{mislove2010you}, which represents the friendship relations between students of Rice University. The friendship graph is an undirected graph that has 1205 nodes with 42443 links between them. Nodes in this graph contain information about students such as student id, age (which is a number between 18 to 22), and major. \vspace{0.3mm} \\  
\textbf{Experimental Setup}: A sub-graph of the \emph{Rice Facebook} dataset is used in the experiments by excluding nodes with a value greater than 20 for their age attribute.
We define two communities in the sub-graph. Nodes with a value of 18 or 19 for their age attribute constitute group $A$  or $\mathcal{V}_A$, and the rest of the nodes form group $B$, $\mathcal{V}_B$. Group $\mathcal{V}_A$ has 97 nodes with 513 intra-connections, while group $\mathcal{V}_B$ has 344 nodes with 7441 intra-connections. There are 1779 inter-connections between nodes of the two communities. We assume an activation probability of 0.01 for 
every link, which is used in the independent cascade model for information propagation.

\subsection{Synthetic Dataset}
\textbf{Dataset Description}: The synthetic dataset is an undirected graph where each node belongs to either of two groups, $\mathcal{V}_A$ and $\mathcal{V}_B$. The size of the groups is set based on a parameter ratio $r$, $0 < r < 1$, where $r\vert \mathcal{V} \vert$ nodes belong to group $\mathcal{V}_A$ and ($(1-r) \vert \mathcal{V} \vert$) nodes belong to group $\mathcal{V}_B$. Nodes are connected based on intra-group, $\mathcal{P}_{intra}^{\mathcal{V}_A}$ or $\mathcal{P}_{intra}^{V_B}$, and inter-group ($\mathcal{P}_{inter}$) connection probabilities. To connect two nodes that belong to the same group (lets say $\mathcal{V}_A$), we do a Bernoulli trial with probability $\mathcal{P}_{intra}^{\mathcal{V}_A}$ and connect the two nodes if the outcome of the trial is 1. Figure \ref{Fig2} shows that the distribution of our Adversarial Graph Embeddings is extremely similar across both groups.
Similarly, two nodes that belong to two different groups are connected only if a Bernoulli trial with probability $\mathcal{P}_{inter}$ is successful. As for the real dataset, there is an activation probability $\mathcal{P}_a$ associated with every link in the network.\\
\textbf{Experimental Setup}: In our experiments, we set $r = 0.3$ which gives 150 nodes in $\mathcal{V}_A$ and 350 nodes in $\mathcal{V}_B$. We use the same intra-group connection probability for both groups, $\mathcal{P}_{intra}^{\mathcal{V}_A} = \mathcal{P}_{intra}^{\mathcal{V}_B} = 0.025$. To ensure two almost-separated communities, we set the inter-group connection probability $\mathcal{P}_{inter} = 0.001$ to be less than the intra-group probability. 
Group $\mathcal{V}_A$ has 134 intra-group connections, while group $\mathcal{V}_B$ has 2843 intra links. There are also 129 inter-group connections between the nodes of the two groups. The activation probability  $\mathcal{P}_a = 0.03$ is the same for all links in the network. 
We tried our method over a range of synthetic data-sets achieving qualitatively similar results. 
Here we include just the results for the above settings.
\section{Results}
\label{Results}
\begin{figure}[h!]

\hspace{1.3cm}
\begin{subfigure}[h!]{0.1\linewidth}
    \centering
    \vfill
    \makebox[0.2\linewidth][c]{\includegraphics[scale=0.16]{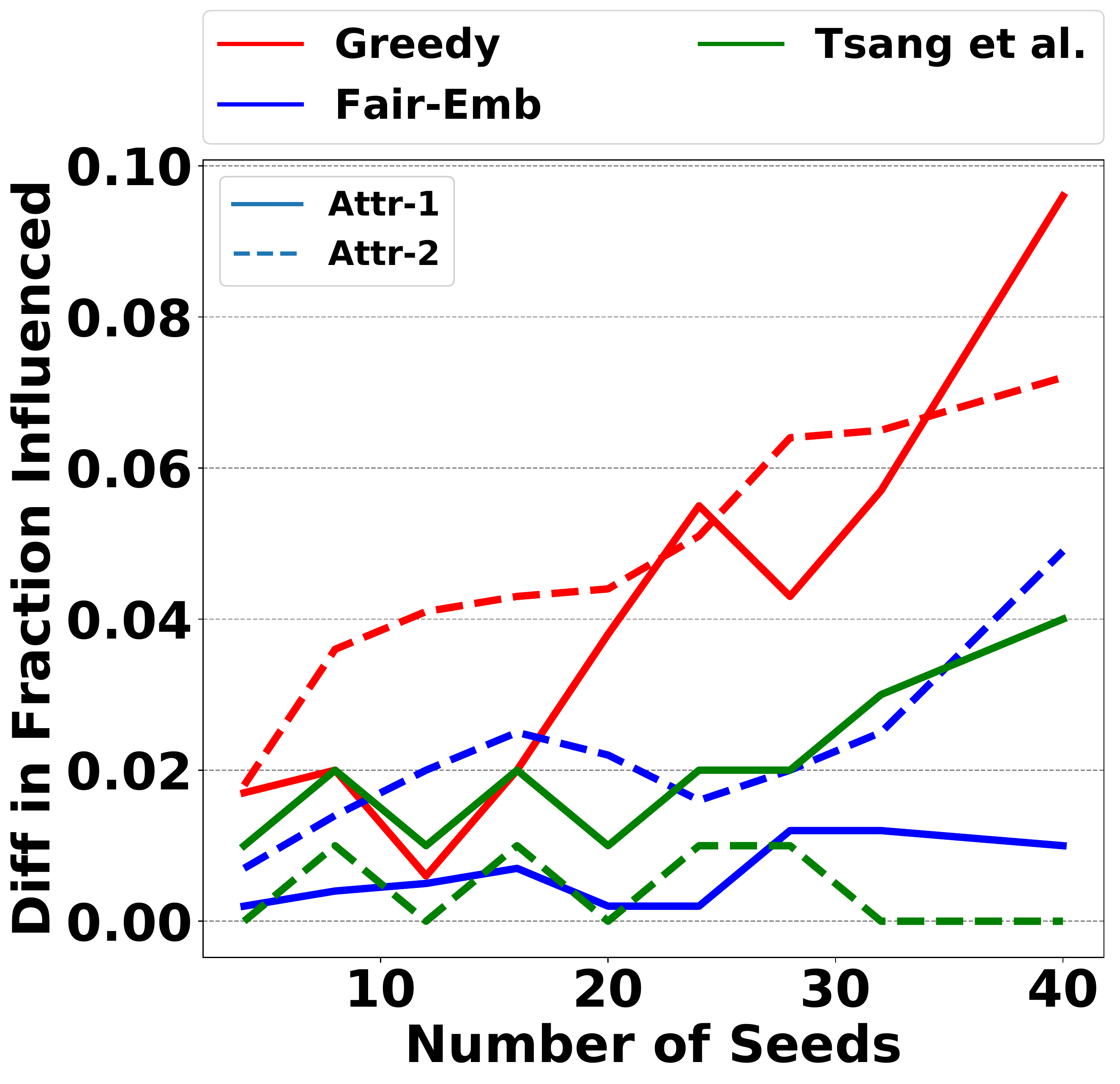}}
   \captionsetup{justification=centering}
   \caption{\small} 
    \label{fig:SynLargePre}
\end{subfigure}
 \hspace{2.5 cm}
\begin{subfigure}[h!]{0.3\linewidth}
    \centering
    \vfill
    \makebox[0.2\linewidth][c]{\includegraphics[scale=0.16]{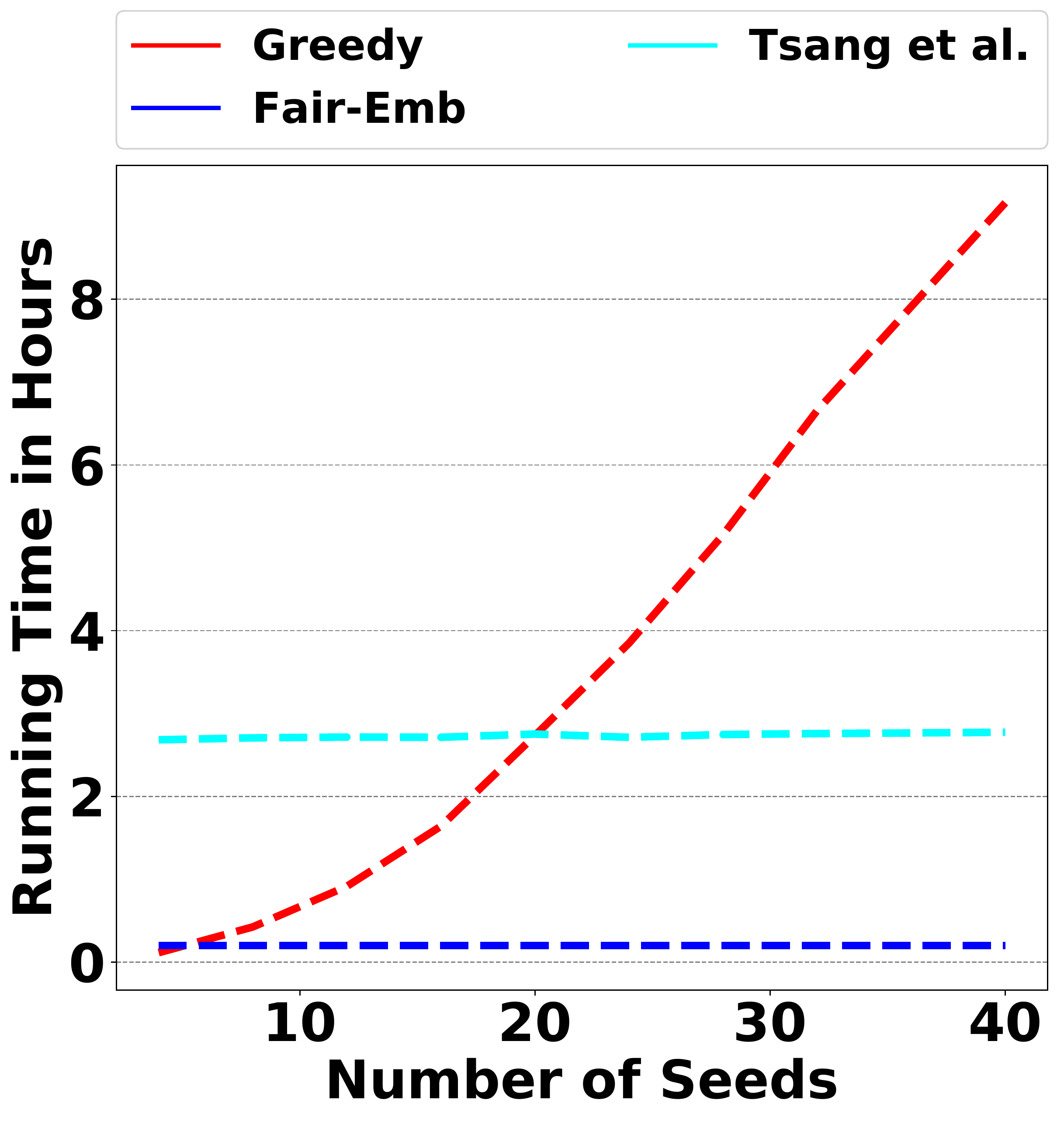}}
   \captionsetup{justification=centering}
    \caption{\small} 
    \label{fig:SynlargeFair}
\end{subfigure}
\vspace{-2.5mm}
\caption{\small (a) The difference in the fraction influenced by the two groups for the case of two sensitive attributes. (b) Comparison of the running time of our proposed method and the baseline.}
\label{Fig:two_attr_res}
\end{figure}

This section includes the results for our proposed method along with the results of two state-of-the-art methods, including \emph{Greedy}  \cite{kempe2003maximizing}, as well as \emph{Tsang et al.} \cite{Tsang_dmt} over synthetic and real data-sets. 
Our experiments follow a two-step process. First, a number of seeds are picked from the input network using our proposed method and the baseline methods.
For the \emph{seed selection},
we utilize the \emph{Fair Selection} approach as it proves to outperform the Normal Selection method. The selected seeds are then passed to an independent cascade model \cite{kempe2003maximizing} to compute the final influenced nodes. 
The results reported show the total fraction of nodes influenced, the fraction of nodes influenced from each community, and the difference between the fractions of nodes influenced from the two communities. Figures \ref{Fig3:Result_real} and \ref{Fig4:Result_synth} show the results over the real and synthetic detests respectively. In the following, we discuss the technical aspects of the methods that appeared in the results and will explain the results in more details.
We conclude the results section by explaining the results for a more general case of two sensitive attributes
(Figure \ref{Fig:two_attr_res}). 

\textbf{Normal Embedding}: Every node of the input network is first described by its row in the adjacency matrix. 
The \emph{Normal Embedding} method learns a low-dimensional representation for each node using an auto-encoder. The auto-encoder ensures that the low-dimensional representation is informative enough to reconstruct the adjacency list by imposing a \emph{cross-entropy} loss that penalizes the model for any dissimilarity between the main and reconstructed adjacency list. 
After getting the embedding, we cluster the points in the embedding space to four clusters (We tested for different numbers of clusters and observed that considering 4 clusters performs best in terms of fairness) and pick the seeds that are nearest to the center of the clusters. This strategy helps to pick nodes with different properties that are likely to influence a good fraction of nodes in the network. Results shows that \emph{Normal Embedding} influences a comparable fraction of the nodes in both the real (Figure \ref{Fig3:Result_real}(a)) and synthetic (Figure \ref{Fig4:Result_synth}(a)) data-sets which is close to the fraction influenced by the \emph{Greedy} method (state-of-the-art classical IM solutions). However, observe that both \emph{Normal Embedding} and \emph{Greedy} are very biased: influencing more nodes from the community of bigger size. In both data-sets, group B is larger and as we see (Figure \ref{Fig3:Result_real}(b, d) and Figure \ref{Fig4:Result_synth}(b, d)), the majority of nodes influenced by the \emph{Normal Embedding} and the \emph{Greedy} methods belong to group B.\\ 

\textbf{Fair Embedding}: To address fairness concerns in \emph{Normal Embedding}, we adversarially train the auto-encoder with a discriminator. Before adversarial training, the discriminator is pre-trained to be able to discriminate between embeddings of the two communities. Through the adversarial training, the auto-encoder tries to learn an embedding for nodes of different groups that are indistinguishable by the discriminator. Using this method, the embedding for nodes of group A will have similar 
distribution to that of group B. 
The adversarial training is followed by a selection algorithm that picks the seeds from the adversarially trained embedding. The selection algorithms are discussed in full details in the \emph{methodology} section.\\
The results show that the \emph{Fair Embedding} method gives a boost to the fraction of nodes influenced from the group of smaller size in both real and synthetic data sets (Fig \ref{Fig3:Result_real}(c, d) and Fig \ref{Fig4:Result_synth}(c, d)). This happens while the total fraction of nodes influenced remains similar (Fig\ref{Fig3:Result_real}(a) and Fig\ref{Fig4:Result_synth}(a)), the \emph{Fair Embedding} only marginally reduces the fraction of nodes influenced from the larger group (See Fig\ref{Fig3:Result_real}(b) and Fig\ref{Fig4:Result_synth}(b)). Overall, the fractions of nodes influenced from the two communities are close to one another 
(Fig\ref{Fig3:Result_real}(c,d) and Fig\ref{Fig4:Result_synth}(c,d)), improving fairness compared to the \emph{Normal Embedding} and  \emph{Greedy} methods. More importantly, in the real dataset, our method significantly outperforms baselines in both maximizing the numbers of people reached and reducing disparity (Figure \ref{Fig3:Result_real}). In synthetic dataset, it seems that our results are very competitive compared to \emph{Tsang et al.} in terms of fairness , however we outperform \emph{Tsang et al.} in terms of the total number of people influenced, and the fraction of people influenced from both $A$ and $B$ communities.

\textbf{Two sensitive attributes}: Figure \ref{fig:SynLargePre} shows the results for the case of two sensitive attributes. As shown, our method outperforms \emph{Greedy} in terms of fairness concerning both sensitive attributes. Our method also shows a comparable performance with \emph{Tsang et al.} in terms of fairness, however, it is more efficient in terms of running time (see Figure \ref{fig:SynlargeFair}).
\section{Conclusion}
In this paper, we proposed a new method to tackle the problem of fair  influence maximization.
We observed that 
existing algorithmic methods to address influence maximization in the population usually lead to considerable disparity.
While there usually exists a trade-off between fairness objectives and maximization objectives proposed, our proposed approach achieves both:
experimental results over synthetic and real-world data-sets demonstrate that our method
is competitive with the prior 
state-of-the-art for maximizing the total number of influenced people, while also significantly decreasing disparity. 

While we demonstrated the effectiveness of our Adversarial Network Embeddings in the context of fair influence maximization, we believe our approach could be used in future work to tackle other fairness tasks over social networks such as fair clustering or fair node-level classification.

\section*{Acknowledgements}
AW acknowledges support from the David MacKay Newton Research Fellowship at Darwin College, The Alan Turing Institute under EPSRC grant EP/N510129/1 \& TU/B/000074, and the Leverhulme Trust via the CFI.

\newpage
{\setstretch{0.9}
\small{
\bibliographystyle{named}
\bibliography{main}}}

\begin{thebibliography}{}

\bibitem[\protect\citeauthoryear{Adel \bgroup \em et al.\egroup }{2019}]{adel}
Tameem Adel, Isabel Valera, Zoubin Ghahramani, and Adrian Weller.
\newblock One-network adversarial fairness.
\newblock In {\em Proceedings of the AAAI Conference on Artificial
  Intelligence}, volume~33, pages 2412--2420, 2019.

\bibitem[\protect\citeauthoryear{Aghaei \bgroup \em et al.\egroup
  }{2019}]{aghaei2019learning}
Sina Aghaei, Mohammad~Javad Azizi, and Phebe Vayanos.
\newblock Learning optimal and fair decision trees for non-discriminative
  decision-making.
\newblock In {\em Proceedings of the AAAI Conference on Artificial
  Intelligence}, volume~33, pages 1418--1426, 2019.

\bibitem[\protect\citeauthoryear{Ali \bgroup \em et al.\egroup
  }{2019}]{Ali2019OnTF}
Junaid Ali, Mahmoudreza Babaei, Abhijnan Chakraborty, Baharan Mirzasoleiman,
  Krishna~P. Gummadi, and Adish Singla.
\newblock On the fairness of time-critical influence maximization in social
  networks.
\newblock {\em ArXiv}, abs/1905.06618, 2019.

\bibitem[\protect\citeauthoryear{Babaei \bgroup \em et al.\egroup
  }{2013}]{babaei2013revenue}
Mahmoudreza Babaei, Baharan Mirzasoleiman, Mahdi Jalili, and Mohammad~Ali
  Safari.
\newblock Revenue maximization in social networks through discounting.
\newblock {\em Social Network Analysis and Mining}, 3(4):1249--1262, 2013.

\bibitem[\protect\citeauthoryear{Babaei \bgroup \em et al.\egroup
  }{2016}]{babaei2016efficiency}
Mahmoudreza Babaei, Przemyslaw Grabowicz, Isabel Valera, Krishna~P Gummadi, and
  Manuel Gomez-Rodriguez.
\newblock On the efficiency of the information networks in social media.
\newblock In {\em Proceedings of the Ninth ACM International Conference on Web
  Search and Data Mining}, pages 83--92, 2016.

\bibitem[\protect\citeauthoryear{Babaei \bgroup \em et al.\egroup
  }{2018}]{babaei2018purple}
Mahmoudreza Babaei, Juhi Kulshrestha, Abhijnan Chakraborty, Fabr{\'\i}cio
  Benevenuto, Krishna~P Gummadi, and Adrian Weller.
\newblock Purple feed: Identifying high consensus news posts on social media.
\newblock In {\em Proceedings of the 2018 AAAI/ACM Conference on AI, Ethics,
  and Society}, pages 10--16. ACM, 2018.

\bibitem[\protect\citeauthoryear{Banerjee \bgroup \em et al.\egroup
  }{2013}]{banerjee2013diffusion}
Abhijit Banerjee, Arun~G Chandrasekhar, Esther Duflo, and Matthew~O Jackson.
\newblock The diffusion of microfinance.
\newblock {\em Science}, 341(6144), 2013.

\bibitem[\protect\citeauthoryear{Benabbou \bgroup \em et al.\egroup
  }{2018}]{benabbou2018diversity}
Nawal Benabbou, Mithun Chakraborty, Xuan-Vinh Ho, Jakub Sliwinski, and Yair
  Zick.
\newblock Diversity constraints in public housing allocation.
\newblock In {\em AAMAS}, pages 973--981, 2018.

\bibitem[\protect\citeauthoryear{Cai \bgroup \em et al.\egroup
  }{2018}]{cai2018comprehensive}
Hongyun Cai, Vincent~W Zheng, and Kevin Chen-Chuan Chang.
\newblock A comprehensive survey of graph embedding: Problems, techniques, and
  applications.
\newblock {\em IEEE Transactions on Knowledge and Data Engineering},
  30(9):1616--1637, 2018.

\bibitem[\protect\citeauthoryear{Cao \bgroup \em et al.\egroup
  }{2015}]{cao2015grarep}
Shaosheng Cao, Wei Lu, and Qiongkai Xu.
\newblock Grarep: Learning graph representations with global structural
  information.
\newblock In {\em Proceedings of the 24th ACM international on conference on
  information and knowledge management}, pages 891--900. ACM, 2015.

\bibitem[\protect\citeauthoryear{Carnes \bgroup \em et al.\egroup
  }{2007}]{carnes2007maximizing}
Tim Carnes, Chandrashekhar Nagarajan, Stefan~M Wild, and Anke Van~Zuylen.
\newblock Maximizing influence in a competitive social network: a follower's
  perspective.
\newblock In {\em EC}, pages 351--360. ACM, 2007.

\bibitem[\protect\citeauthoryear{Fish \bgroup \em et al.\egroup
  }{2019}]{fish_dmt}
Benjamin Fish, Ashkan Bashardoust, danah boyd, Sorelle~A. Friedler, Carlos
  Scheidegger, and Suresh Venkatasubramanian.
\newblock {Gaps in Information Access in Social Networks}.
\newblock In {\em {WWW}}, 2019.

\bibitem[\protect\citeauthoryear{Gomez-Rodriguez \bgroup \em et al.\egroup
  }{2010}]{leskovec2010inferring}
M.~Gomez-Rodriguez, J.~Leskovec, and A.~Krause.
\newblock {Inferring Networks of Diffusion and Influence}.
\newblock In {\em KDD}, 2010.

\bibitem[\protect\citeauthoryear{Goodfellow \bgroup \em et al.\egroup
  }{2014}]{goodfellow2014generative}
Ian Goodfellow, Jean Pouget-Abadie, Mehdi Mirza, Bing Xu, David Warde-Farley,
  Sherjil Ozair, Aaron Courville, and Yoshua Bengio.
\newblock Generative adversarial nets.
\newblock In {\em Advances in neural information processing systems}, pages
  2672--2680, 2014.

\bibitem[\protect\citeauthoryear{Goyal \bgroup \em et al.\egroup
  }{2013}]{goyal2013minimizing}
Amit Goyal, Francesco Bonchi, Laks~VS Lakshmanan, and Suresh
  Venkatasubramanian.
\newblock On minimizing budget and time in influence propagation over social
  networks.
\newblock {\em Social network analysis and mining}, 3(2):179--192, 2013.

\bibitem[\protect\citeauthoryear{Grover and
  Leskovec}{2016}]{grover2016node2vec}
Aditya Grover and Jure Leskovec.
\newblock node2vec: Scalable feature learning for networks.
\newblock In {\em Proceedings of the 22nd ACM SIGKDD international conference
  on Knowledge discovery and data mining}, pages 855--864. ACM, 2016.

\bibitem[\protect\citeauthoryear{Hamilton \bgroup \em et al.\egroup
  }{2017a}]{hamilton2017inductive}
Will Hamilton, Zhitao Ying, and Jure Leskovec.
\newblock Inductive representation learning on large graphs.
\newblock In {\em Advances in Neural Information Processing Systems}, pages
  1024--1034, 2017.

\bibitem[\protect\citeauthoryear{Hamilton \bgroup \em et al.\egroup
  }{2017b}]{hamilton2017representation}
William~L Hamilton, Rex Ying, and Jure Leskovec.
\newblock Representation learning on graphs: Methods and applications.
\newblock {\em arXiv preprint arXiv:1709.05584}, 2017.

\bibitem[\protect\citeauthoryear{Keikha \bgroup \em et al.\egroup
  }{2020}]{keikha2020influence}
Mohammad~Mehdi Keikha, Maseud Rahgozar, Masoud Asadpour, and Mohammad~Faghih
  Abdollahi.
\newblock Influence maximization across heterogeneous interconnected networks
  based on deep learning.
\newblock {\em Expert Systems with Applications}, 140:112905, 2020.

\bibitem[\protect\citeauthoryear{Kempe \bgroup \em et al.\egroup
  }{2003}]{kempe2003maximizing}
David Kempe, Jon Kleinberg, and {\'E}va Tardos.
\newblock Maximizing the spread of influence through a social network.
\newblock In {\em KDD}, 2003.

\bibitem[\protect\citeauthoryear{Khajehnejad}{2019}]{khajehnejad2019simnet}
Moein Khajehnejad.
\newblock Simnet: Similarity-based network embeddings with mean commute time.
\newblock {\em PloS one}, 14(8):e0221172, 2019.

\bibitem[\protect\citeauthoryear{Kourtellis \bgroup \em et al.\egroup
  }{2013}]{kourtellis2013identifying}
Nicolas Kourtellis, Tharaka Alahakoon, Ramanuja Simha, Adriana Iamnitchi, and
  Rahul Tripathi.
\newblock Identifying high betweenness centrality nodes in large social
  networks.
\newblock {\em Social Network Analysis and Mining}, 3(4):899--914, 2013.

\bibitem[\protect\citeauthoryear{Leskovec \bgroup \em et al.\egroup
  }{2007}]{leskovec2007cost}
Jure Leskovec, Andreas Krause, Carlos Guestrin, Christos Faloutsos, Jeanne
  VanBriesen, and Natalie Glance.
\newblock Cost-effective outbreak detection in networks.
\newblock In {\em KDD}, pages 420--429. ACM, 2007.

\bibitem[\protect\citeauthoryear{Li \bgroup \em et al.\egroup
  }{2018}]{li2018influence}
Yuchen Li, Ju~Fan, Yanhao Wang, and Kian-Lee Tan.
\newblock Influence maximization on social graphs: A survey.
\newblock {\em TKDE}, 30(10):1852--1872, 2018.

\bibitem[\protect\citeauthoryear{Madras \bgroup \em et al.\egroup
  }{2018}]{madras}
David Madras, Elliot Creager, Toniann Pitassi, and Richard Zemel.
\newblock Learning adversarially fair and transferable representations.
\newblock In {\em International Conference on Machine Learning}, pages
  3384--3393, 2018.

\bibitem[\protect\citeauthoryear{Mirzasoleiman \bgroup \em et al.\egroup
  }{2012}]{mirzasoleiman2012immunizing}
Baharan Mirzasoleiman, Mahmoudreza Babaei, and Mahdi Jalili.
\newblock Immunizing complex networks with limited budget.
\newblock {\em EPL (Europhysics Letters)}, 98(3):38004, 2012.

\bibitem[\protect\citeauthoryear{Mislove \bgroup \em et al.\egroup
  }{2010}]{mislove2010you}
Alan Mislove, Bimal Viswanath, Krishna~P Gummadi, and Peter Druschel.
\newblock You are who you know: inferring user profiles in online social
  networks.
\newblock In {\em WSDM}, pages 251--260. ACM, 2010.

\bibitem[\protect\citeauthoryear{Richardson and
  Domingos}{2002}]{richardson2002mining}
Matthew Richardson and Pedro Domingos.
\newblock Mining knowledge-sharing sites for viral marketing.
\newblock In {\em KDD}, 2002.

\bibitem[\protect\citeauthoryear{Tang \bgroup \em et al.\egroup
  }{2015}]{tang2015line}
Jian Tang, Meng Qu, Mingzhe Wang, Ming Zhang, Jun Yan, and Qiaozhu Mei.
\newblock Line: Large-scale information network embedding.
\newblock In {\em Proceedings of the 24th international conference on world
  wide web}, pages 1067--1077. International World Wide Web Conferences
  Steering Committee, 2015.

\bibitem[\protect\citeauthoryear{Tsang \bgroup \em et al.\egroup
  }{2019}]{Tsang_dmt}
Alan Tsang, Bryan Wilder, Eric Rice, Milind Tambe, and Yair Zick.
\newblock {Group-Fairness in Influence Maximization}.
\newblock {\em arXiv preprint arXiv:1903.00967}, 2019.

\bibitem[\protect\citeauthoryear{Wang \bgroup \em et al.\egroup
  }{2016}]{wang2016structural}
Daixin Wang, Peng Cui, and Wenwu Zhu.
\newblock Structural deep network embedding.
\newblock In {\em Proceedings of the 22nd ACM SIGKDD international conference
  on Knowledge discovery and data mining}, pages 1225--1234. ACM, 2016.

\bibitem[\protect\citeauthoryear{Ye \bgroup \em et al.\egroup
  }{2012}]{ye2012exploring}
Mao Ye, Xingjie Liu, and Wang-Chien Lee.
\newblock Exploring social influence for recommendation: a generative model
  approach.
\newblock In {\em SIGIR}, pages 671--680. ACM, 2012.

\end{thebibliography}
\bigskip

\end{document}